\documentclass[10pt,twocolumn,letterpaper]{article}

\usepackage{cvm}
\usepackage{times}
\usepackage{epsfig}
\usepackage{graphicx}
\usepackage{amsmath}
\usepackage{amssymb}
\usepackage{textcomp}
\usepackage{stfloats}
\usepackage{url}
\usepackage{verbatim}
\usepackage{graphicx}
\usepackage{cite}
\usepackage{multirow}
\usepackage{threeparttable}
\usepackage{booktabs}
\usepackage{color} 
\usepackage[misc]{ifsym}

\usepackage{amsmath}
\usepackage{diagbox}

\usepackage[pagebackref=true,breaklinks=true,letterpaper=true,colorlinks,bookmarks=false]{hyperref}

\cvmfinalcopy

\def\cvmPaperID{****} 

\ifcvmfinal\fi
\begin{document}

\title{Appearance Blur-driven AutoEncoder and Motion-guided Memory Module for Video Anomaly Detection}

\author{Jiahao Lyu, Minghua Zhao$^{(\textrm{\Letter})}$, Jing Hu, Xuewen Huang, Shuangli Du, Cheng Shi, Zhiyong Lv\\
School of Computer Science and Engineering, Xi'an University of Technology\\
Xi'an, China\\
{\tt\small zhaominghua@xaut.edu.cn}
}

\maketitle

\begin{abstract}
	Video anomaly detection (VAD) often learns the distribution of normal samples and detects the anomaly through measuring significant deviations, but the undesired generalization may reconstruct a few anomalies thus suppressing the deviations. Meanwhile, most VADs cannot cope with cross-dataset validation for new target domains, and few-shot methods must laboriously rely on model-tuning from the target domain to complete domain adaptation. To address these problems, we propose a novel VAD method with a motion-guided memory module to achieve cross-dataset validation with zero-shot. First, we add Gaussian blur to the raw appearance images, thereby constructing the global pseudo-anomaly, which serves as the input to the network. Then, we propose multi-scale residual channel attention to deblur the pseudo-anomaly in normal samples. Next, memory items are obtained by recording the motion features in the training phase, which are used to retrieve the motion features from the raw information in the testing phase. Lastly, our method can ignore the blurred real anomaly through attention and rely on motion memory items to increase the normality gap between normal and abnormal motion. Extensive experiments on three benchmark datasets demonstrate the effectiveness of the proposed method. Compared with cross-domain methods, our method achieves state-of-the-art performance without adaptation during testing. 
\end{abstract}

\section{Introduction}

\begin{figure}[ht]
	\centering  
	\includegraphics[width=\linewidth]{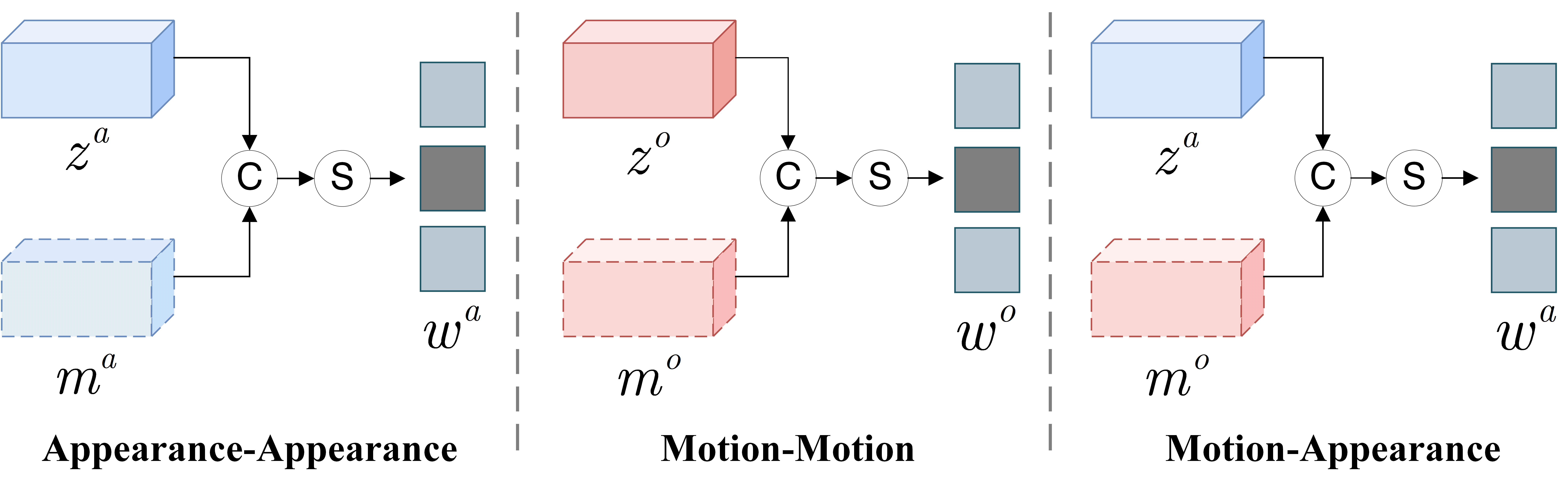}
	\caption{Typical solutions of memory modules in video anomaly detection. \emph{Left:} Most memory modules reconstruct entire appearance features, but are limited by the memory item size. \emph{Middle:} A few memory modules reconstruct motion features, but are limited to RoI bounding boxes. \emph{Right:} The new memory module proposed is not limited by the above shortcomings. By retrieving background-independent motion features, it is simple to implement VAD and cross-domain detection.}
	\label{fig:f1}
\end{figure}
Video surveillance devices are extensively employed in transportation and industry, undeniably augmenting the burden of manual detection, and potentially leading to overlooked anomalies. Hence, video anomaly detection (VAD) has been introduced as a crucial technology into intelligent surveillance systems. With surging surveillance videos, labeling samples becomes particularly tedious \cite{yu2022deep}. Fully-unsupervised anomaly detection methods \cite{yu2022deep, pang2020self} remain in their infancy, and their performances depend on the imbalance of samples in the dataset and the differences between the normal and anomaly \cite{liu2023generalized}. Therefore, one-class classification (OCC) unsupervised\footnote{To avoid ambiguity, in this work, we define OCC fall in the category of unsupervised.} methods have been considered as the predominant method for solving VAD problems \cite{pang2021deep}.

Most OCC methods fall into two categories: prediction-based and reconstruction-based: (1) prediction-based methods \cite{liu2018future} predict future frame by inputting multiple frames, and identify anomalies by measuring the deviations between the predicted and the real frame. (2) reconstruction-based methods \cite{gong2019memorizing} identify anomalies by reconstructing the inputs and measuring the reconstruction errors. OCC primarily uses generative models as benchmarks, such as autoencoder (AE) \cite{liu2023msn} or generative adversarial network (GAN) \cite{li2023multi}. These methods can be further categorized into single-stream \cite{liu2023msn,zhang2022hybrid,chang2022video} or dual-stream \cite{wang2023memory,zhong2022bidirectional,fang2020anomaly} codec architectures. Besides decoupling appearance features, motion features are used as auxiliary information in most dual-stream networks. Among them, the optical flow-based methods have achieved outstanding performance. However, the shortcoming of optical flow is its high computational complexity \cite{ilg2017flownet}, which limits the detection efficiency. Furthermore, the memory modules \cite{gong2019memorizing,park2020learning} have been adopted as a normality discrimination way.


Existing OCCs have mainly been modeling the entire raw image and completing anomaly detection by decoupling various normal samples \cite{park2020learning}. These methods work well for handling anomalies with significant deviations, but face challenges when the boundaries between the normal and anomaly are ambiguous. Furthermore, these methods have limited performance in cross-domain VAD problems. Therefore, the existing methods have two main shortcomings: \textbf{(1)} Existing methods aim to enhance model sensitivity to anomalies by reconstructing the entire video frame and introducing various operations, such as the attention mechanism \cite{zhang2022hybrid,le2023attention,wang2023video,kommanduri2024dast,singh2024attention}, multi-scale feature extraction \cite{liu2023msn,zhong2022bidirectional}, and memory module \cite{gong2019memorizing,liu2023msn,park2020learning,cai2021appearance}. These methods are inevitably influenced by the negative impact of complex and redundant background information. \textbf{(2)} Most previous methods do not consider the cross-domain VAD problem and can only achieve good performance in the source domain. Few-shot methods \cite{lu2020few, lv2021learning, wang2022few, huang2022boosting,zhang2024cognition}, meanwhile, must undergo laborious target domain fine-tuning to achieve the adaptation. Moreover, the memory module-based methods record the background information of the source domain and further hinder cross-domain adaptation.

To this end, the motivation of our work is to focus on motion features that may contain anomalies. We achieve this by using memory items to record motion features while ignoring the detrimental effect of the background. Figure \ref{fig:f1} illustrates different memory modules in VAD. Focusing on clean motion features means that the proposed memory module can be independent from the background, and can distinguish between normal and abnormal motion. Moreover, from the visualization results of most VADs, it can be observed that the anomalies are presented as blurred, so we consider the VAD process as adding blur only to the anomalies, and thus we transform the VAD into an inverse process of the above process, i.e., we complete the deblurring process for the blurred normal features, and the anomaly features continue to retain the blurred features. Therefore, we design a skip connection with multi-scale residual channel attention (MRCA) to utilize appearance images with Gaussian blur as the pseudo-anomaly. In the training phase, the MRCA ensures the model removes the blur from normal samples. In the testing phase, the blurred real anomaly is ignored by the attention. Finally, our work achieves efficient anomaly detection and addresses the cross-domain VAD. Our contributions are summarized below:
\begin{itemize}
	\item  We transform VAD from a fixed decoupling method of learning normal images into the process of eliminating Gaussian blur pseudo-anomaly, enhancing the robustness of the model to anomaly detection.
	\item  We focus on motion features in videos and propose a motion-guided memory module. The normal motion distribution is learned in the training phase, and motion memory items are used to retrieve motion features from appearance information in the testing phase.
	\item  Extensive experiments on the three benchmark datasets show that our method has advantages against state-of-the-art methods. Importantly, the cross-dataset validations demonstrate the scene generalizability of the proposed method.

\end{itemize}

\begin{figure*}[ht]
	\centering
	\includegraphics[width=\linewidth]{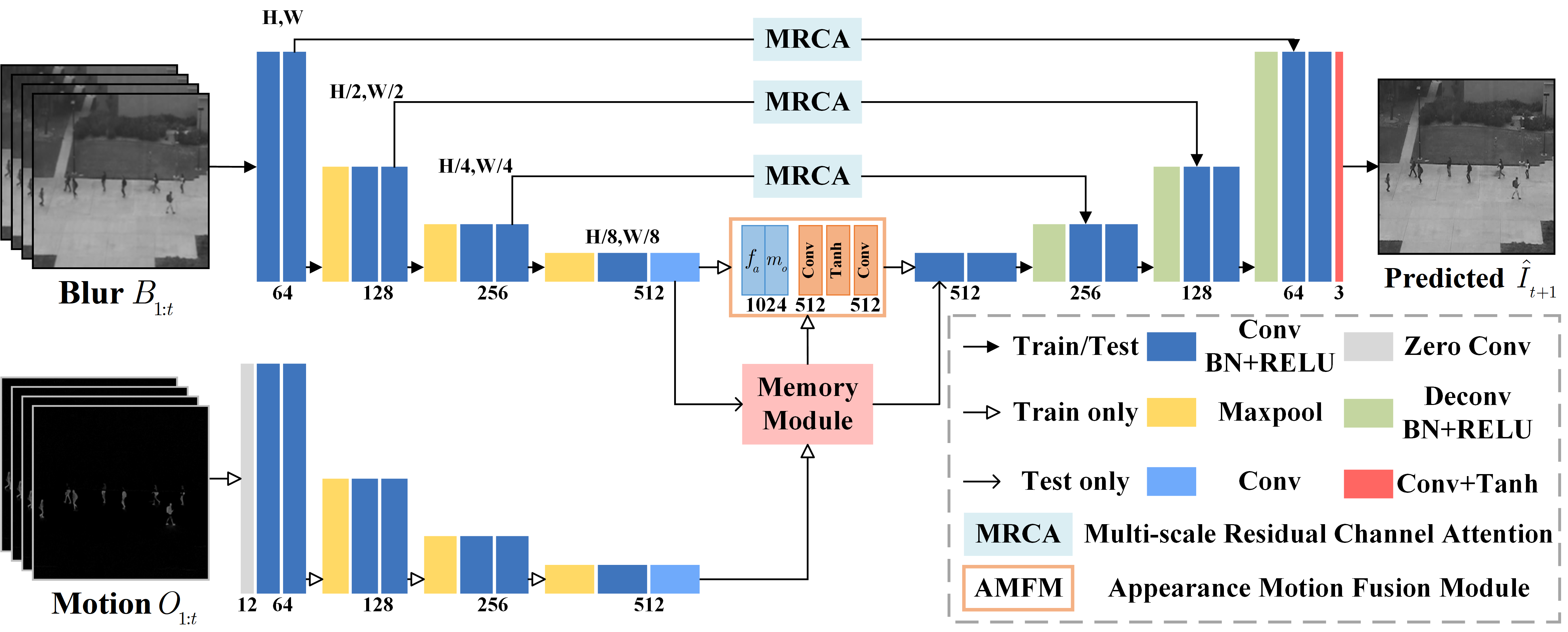}
	\caption{Overview framework of the proposed method. It employs a dual-stream AE with input Gaussian blur appearance images $B_{1:t}$ and motion images $O_{1:t}$ to output a predicted image $\hat{I}_{t+1}$, and consists of skip connections with MRCA, a motion-guided memory module, an appearance motion fusion module. During testing, just input blurred appearance images. The horizontal dimension indicates the number of output channels. H and W denote the height and width of features, respectively.}
	\label{fig:f2}
	
\end{figure*}

%
%

\section{Related work}
\subsection{Unsupervised VAD}
Unsupervised VAD \cite{cheng2023spatial,yang2023video} learns a model of normality using solely normal samples in the training phase and identifies the anomaly samples by measuring deviations during the testing phase. Many works are based on AE to complete the reconstruction or prediction of the entire raw image. Zhang et al. \cite{zhang2022hybrid} proposed a hybrid attention and motion constrained anomaly detection method, in which channel attention and spatial attention are added in the encoding and skip connection stages, respectively. This allows the model to focus on normal samples by attention. Wang et al. \cite{wang2021robust} proposed a prediction network based on multi-path ConvGRU, which processes feature information at different scales while obtaining the temporal relationship between consecutive frames, thereby reducing the attention given to static and background features. In addition, several works are developed on GAN \cite{liu2018future,li2023multi,singh2024attention}. These methods rely on stronger feature expression capabilities and achieve better performance.

Meanwhile, some new variants of detection methods are available. Zhong et al. \cite{zhong2022bidirectional} proposed a bidirectional frame prediction method to generate video frames at both ends of the video sequence through forward and backward prediction. Moreover, they designed a new evaluation method to achieve multi-scale anomaly evaluation using an error pyramid and mean pooling to detect target objects of different sizes effectively. Fang et al. \cite{fang2020anomaly} proposed another bidirectional prediction method that utilizes the video frames at both ends to predict the intermediate frames simultaneously and performs anomaly detection on the two results based on consistency evaluation standards. Yang et al. \cite{yang2023video} proposed a prediction-like method based on keyframe recovery to recover complex motion features in videos through cross-attention. Shi et al. \cite{shi2023video} implemented a hybrid model for VAD by ranking a combination of three different tasks: prediction, reconstruction, and classification. 

\subsection{Cross-domain Detection}

Mainstream unsupervised VADs fail to consider cross-domain validation \cite{liu2023generalized}, can usually achieve good performance only in the source domain, and cannot be effectively transferred from the source domain to another domain. The real world always contains multiple dissimilar scenes, and how to consider cross-domain detection under scenario differences is a significant challenge for VAD. To address this, some works\cite{lu2020few, lv2021learning, wang2022few, huang2022boosting} introduce meta-learning methods to explore the potential for solving the challenge. For instance, Lu et al. \cite{lu2020few} based on the meta-learning method, first introduce few-shot scene-adaptive, which effectively adapts to new domains by using only a few frames from the relative scenario. Moreover, Aich et al. \cite{aich2023cross} perform pseudo-anomaly synthesis using untrained CNNs with auxiliary Data. The proposed Normalcy Classifier Module can discriminate the differences between real and pseudo anomalies, which not only reduces the overfitting of normal video features in the source domain, but also achieves cross-domain detection by discriminating different anomalies in different target domains from the extremely diverse pseudo-anomalies.

\subsection{Memory Module}

Memory modules are widely used in video prediction \cite{hu2023dynamic}, video segmentation \cite{lee2023unsupervised}, industrial anomaly detection \cite{xing2023visual}, and VAD \cite{wang2023memory}. Most VADs have completed normality judgment by designing various memory modules \cite{wang2023memory,gong2019memorizing,park2020learning,cai2021appearance}, and part of the work focuses on how to achieve the cooperation and collaboration of multiple memory modules \cite{liu2021hybrid,liu2023msn} to achieve the discrimination of different dimensional features. For instance, MemAE \cite{gong2019memorizing} is the memory module with Hard Shrinkage, which alleviates the possibility that a few anomalies may remain in the memory items. MNAD \cite{park2020learning} is the memory module that includes two modes Read and Update. Especially in the test phase, weighted conventional scores are used to prevent memory items from recording abnormal information. Liu et al. \cite{liu2021hybrid} explored three forms of multi-memory module cooperation, and finally inserted memory modules to complete optical flow reconstruction when decoding different features. Unlike the work \cite{liu2021hybrid}, MSN-net \cite{liu2023msn} implemented the anomaly detection of multi-scale features by using three individual memory modules during skip connections. 

Most modules belong to \emph{Left} in Figure \ref{fig:f1}, which aims to reconstruct entire appearance features through stored memory items. However, if the memory item is too small, it will easily lose its normal distribution, and if it is too large, it will help the anomaly to be reverted. And \cite{liu2021hybrid} belongs to \emph{Middle} in Figure \ref{fig:f1}. Its purpose is to reconstruct the optical flow features, as a condition to assist conditional variational autoencoder in making final predictions.

\subsection{Pseudo-anomaly Data}

Unsupervised VAD focuses on the raw appearance, but the difference between normal and anomaly is only in motion features, the redundant background increases the burden on the model training \cite{morais2019learning} and causes over-fitting for the anomaly. Therefore, some works proposed using the pseudo-anomaly \cite{astrid2023pseudobound,aich2023cross} or random mask \cite{zhong2022cascade} to improve the robustness of the model. Zhong et al. \cite{zhong2022cascade} multiplied the mask patch composed of 0 or 1 with the input frame pixel by pixel, and video frames with black pixel blocks are treated as the pseudo-anomaly. Astrid et al. \cite{astrid2023pseudobound} employed five ways to composite various pseudo-anomaly, including skip frames, add noise, random fusion frames, etc. Based on GAN, Li et al. \cite{li2023multi} utilized an initially poorly-performing generator to create low-quality predicted frames as pseudo-anomaly, thereby training the discriminator to better distinguish the anomaly. Zhu et al. \cite{zhu2023cross} proposed the multi-cross illumination datasets from the perspective of video illumination to make the baseline model more accurate and robust. The aforementioned methods alleviated over-fitting in unsupervised VAD from the perspective of data augmentation. Still, they never considered the differences in morphology and appearance between normal and abnormal after reconstruction or prediction.

\section{METHODOLOGY}

According to the above descriptions, how to avoid undesired anomaly generalization in target domains and simultaneously cope with cross-domain validation has significant gaps. Therefore, we introduce a novel VAD method based on the Gaussian blur pseudo-anomaly and motion-guided memory module. Figure \ref{fig:f2} shows the framework of our proposed method. Different from previous work, the proposed method needs extra auxiliary motion images in the training phase and no longer needs them in the testing phase, which guarantees fast inference speed without complex computation.



The basic structure of this work is presented in Section \ref{sec:sc3.1}, including how to construct the dual-stream AE, the attention mechanism, and the feature fusion module, as well as the application of zero-convolution to the motion encoder. Section \ref{sec:sc3.2} describes applying the motion-guided memory module to the training and testing phase, respectively. We elaborate on the loss functions in training and describe the details of anomaly detection during testing in Section \ref{sec:sc3.3}.

\subsection{Dual-stream Network Architecture} \label{sec:sc3.1}
\subsubsection{Gaussian blur-driven AE} 

Based on U-net \cite{ronneberger2015u}, we design a dual-encoder and single-decoder AE, to capture the representation of the blurred image $B$ in the temporal and spatial domains, and to complete anomaly detection. Typically, unsupervised methods are trained using raw images $I$, and can easily recognize anomalies with remarkable differences, but generalization may reconstruct a few anomalies. Moreover, we observed that in most VADs \cite{yu2020cloze,yang2023video,liu2023msn,georgescu2021background}, when anomalies are generated through reconstruction or prediction, anomalies are eventually represented as blurred patterns. Therefore, we have transformed the original VAD mode into the Gaussian blur elimination process, and the Gaussian blur pseudo-anomaly images $B$ can be formulated as follows:

\begin{equation}\label{eq:1}
B=I \otimes G(k,\sigma),
\end{equation}
where $\otimes$ denotes element-wise multiplication, $G(\cdot)$ is the Gaussian blur filter, $k$ and $\sigma$ are the Gaussian kernel size and standard deviation, respectively. 

Notice that the blurring process requires only $10^{-3}$ seconds to blur a $256\times256$ image. For example, in the same framework, blurring of all images for an epoch just requires an auxiliary 5s (from 158.8s to 163.9s). The auxiliary 5s is considered to be negligible, when compared with the 158.8s.

\subsubsection{Multi-scale residual channel attention} 

Attention mechanism is widely used in many vision tasks, and attention with residual units can further enhance the performance of tasks \cite{wang2017residual}. Meanwhile, to avoid copying abnormal features in skip connections, we propose the skip connection with MRCA to help the model focus on normal pseudo-anomaly and ignore the blurred real anomaly. 

Figure \ref{fig:f3} shows MRCA, it taking feature maps $F\in\mathbb{R}^{C\times H\times W}$ as input. Firstly, the shallow intermediate feature map $F^{z}$ is obtained through two different two-dimensional (2D) convolutions. Secondly, we use element-wise multiplication between $F^{z}$ and the channel attention map $M\in\mathbb{R}^{C\times 1\times 1}$ to obtain ${F}'$. Finally, the upper and lower attention maps are added with $F$ to obtain the final refined output ${F}''$. The overall MRCA can be as follows:

\begin{figure}[ht]
	\centering
	\includegraphics[width=\linewidth]{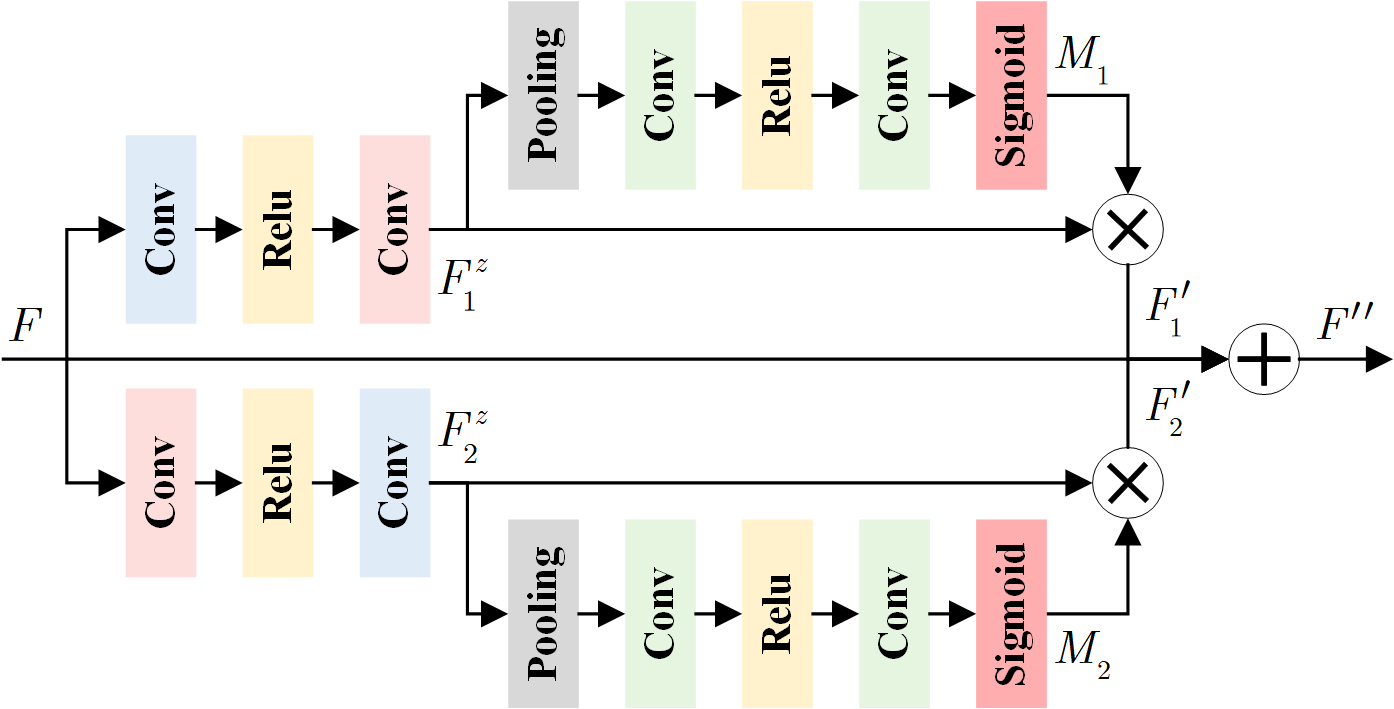}
	\caption{Multi-scale residual channel attention.}
	\label{fig:f3}
\end{figure}

\begin{equation}\label{eq:2}
\begin{gathered}
F^{\prime \prime}=F_{1}^{\prime}+F_{2}^{\prime}+F,
\end{gathered}
\end{equation}

\begin{equation}\label{eq:2.1}
\begin{gathered}
F_{i}^{\prime}=M_{i}\left(F_{i}^{z}\right) \otimes F_{i}^{z}, \forall i\in \left \{ 1,2 \right \}.
\end{gathered}
\end{equation}

The shallow features $F_{1}^{z}$ and $F_{2}^{z}$ can be expressed as:

\begin{equation}\label{eq:3}
\begin{gathered}
F_{1}^{z}=W_{1} \delta\left(W_{2} F\right),
\end{gathered}
\end{equation}

\begin{equation}\label{eq:3.1}
\begin{gathered}
F_{2}^{z}=W_{2} \delta\left(W_{1} F\right),
\end{gathered}
\end{equation}
where $\delta$ is the ReLU activation function, $W_{1}\in \mathbb{R}^{C\times C}$ and $W_{2}\in \mathbb{R}^{C\times C}$ are 2D convolutions with different filter sizes. The filter sizes of $W_{1}$ and $W_{2}$ for different dimensions and datasets are described in Section \ref{sec:sc4.1}. We use the global average-pooling to obtain the attention feature map $M$ of $F^{z}$:
\begin{equation}\label{eq:4}
M=\xi (W_{3}\delta (W_{4}P_{avg}(F^{z}))),
\end{equation}
where $\xi $ the Sigmoid activation function, $W_{3}\in \mathbb{R}^{C/r\times C}$ and $W_{4}\in \mathbb{R}^{C\times C/r}$ are 2D convolutions kernels whose size is $1 \times 1$, and $P_{avg}$ is the global average-pooling operation.

\subsubsection{Zero convolution removes motion noise} 

In contrast to building pseudo-anomaly in appearance images, we want to record motion distributions from the cleanest motion images. Foreground extraction \cite{huynh2016nic} or optical flow estimation networks \cite{ilg2017flownet} are unable to achieve the desired effect, namely, it cannot avoid noise. Therefore, we add zero convolutions \cite{zhang2023adding} to the motion encoder, and the rest of the structure is the same as the appearance encoder. Zero convolution has recently been used in diffusion models \cite{zhang2023adding}, which ensures that harmful noise does not damage the model fine-tuning of trainable copy. Therefore, to pursue cleaner motion features, we use zero convolution to let the network parameters learn the feature distribution from scratch, thereby reducing the harmful noise that may be hidden in different motion features.

\subsubsection{Appearance motion fusion module} 

We fuse and concatenate the appearance features $f_{a}$ from the appearance encoder and the motion features $m_{o}$ obtained through the motion memory item as a new representation for the decoder. Since the appearance motion fusion module (AMFM) is just used in the training phase, and the testing phase consists only of the appearance encoder, to avoid destroying the feature distribution during testing, the AMFM consists of two convolutions with the filter size of $1 \times 1$ and a Tanh activation function.

\subsection{Motion-guided Memory Module}\label{sec:sc3.2}

\begin{figure}[ht]
	\centering
	\includegraphics[width=\linewidth]{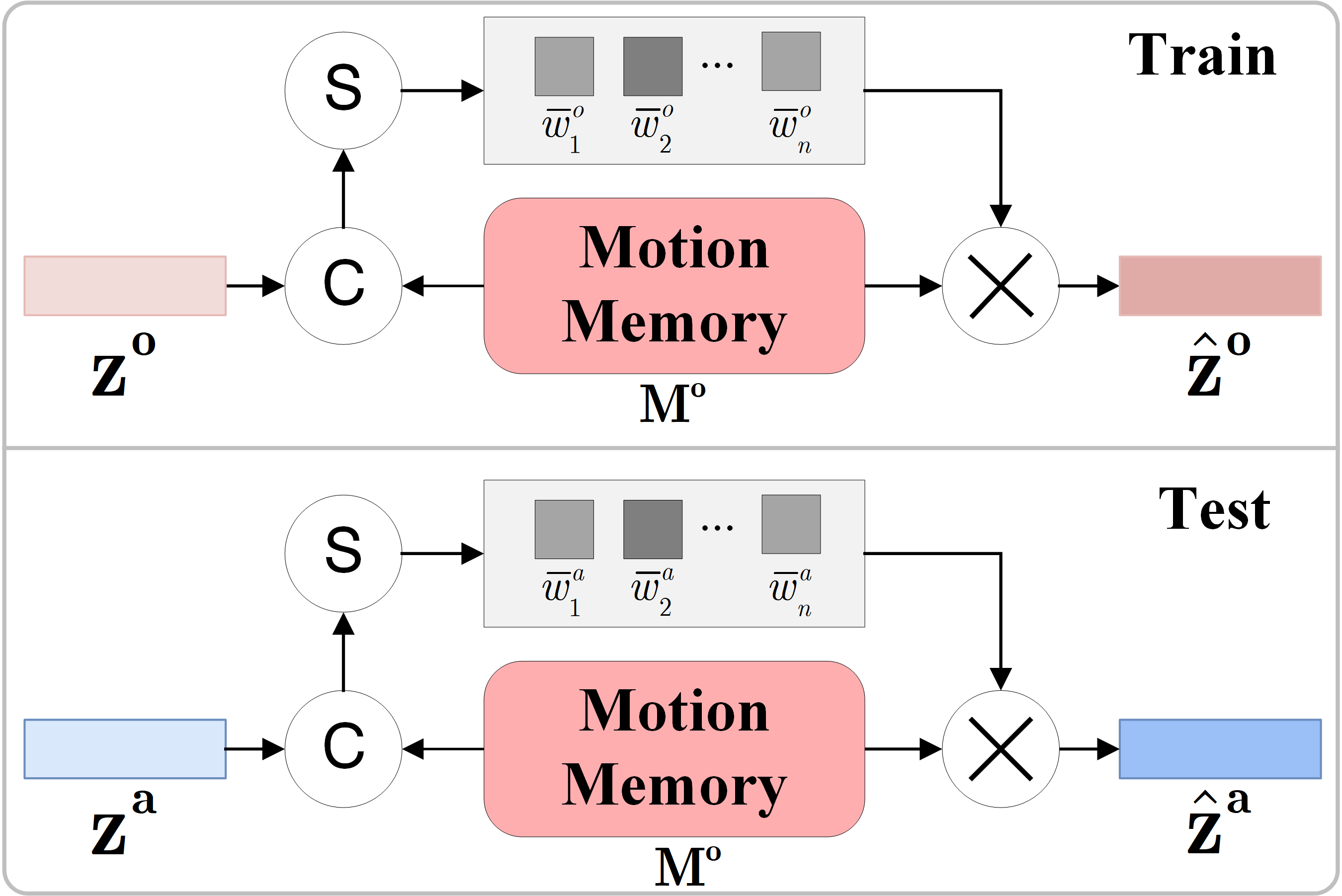}
	\caption{Motion-guided memory module. c: Cosine similarities, s: Softmax function. See text for details.}
	\label{fig:f4}
\end{figure}

The new motion-guided memory module, as shown in Figure \ref{fig:f4}, aims to record normal motion distribution from clean motion feature representation so that it extracts normal motion features from appearance information during the test phase, excluding possible abnormal motion. 

First, during the training phase, the motion feature map $F^{o}\in \mathbb{R}^{C\times H\times W}$ obtained from the motion encoder is expanded along the channel dimension into motion query maps $\mathbf{z^{o}} \in \mathbb{R}^{C}$ and input into the motion memory pool. The motion memory pool is a 2D matrix $\mathbf{M^{o}} \in \mathbb{R}^{N \times C} $, where $N$ represents the number of memory items, and $C$ is the dimension of each memory item $m_{i}^{o}(i=1, \ldots, N)$. Then, $m^{o}_{i}$ records the normal motion feature. Finally, calculate the similarity between $\mathbf{z^{o}}$ and $m^{o}_{i}$:

\begin{equation}\label{eq:5}
w_{i}^{o}=\frac{\mathbf{z^{o}}{(m_{i}^{o})}^{T}}{\left \|\mathbf{z^{o}}  \right \| \left \|m_{i}^{o}  \right \| }.
\end{equation}

Then we execute the Softmax function for each $w_{i}^{o}$ to calculate the weight $\bar{w} _{i}^{o}$ as follows:

\begin{equation}\label{eq:6}
\bar{w}_{i}^{o}=\frac{exp(w_{i}^{o})}{ {\textstyle \sum_{j=1}^{N}exp(w_{j}^{o})} }.
\end{equation}

The final reconstructed motion features $\mathbf{\hat{z}^{o}}$ are obtained by retrieving the most similar memory item to $\mathbf{z^{o}}$ in the memory module:

\begin{equation}\label{eq:7}
\mathbf{\hat{z}^{o}}=\bar{\mathbf{w}}^\mathbf{o}\mathbf{M^{o}}= {\textstyle \sum_{i=1}^{N}\bar{w}_{i}^{o}m_{i}^{o}}.
\end{equation}

In the testing phase, we input the appearance query map $\mathbf{z^{a}} \in \mathbb{R}^{C}$ composed of the appearance feature map $F^{a}\in \mathbb{R}^{C\times H\times W}$, and reconstruct the normal motion in the appearance features through $\mathbf{M^{o}}$. Specifically, using Eq. (\ref{eq:5}) and Eq. (\ref{eq:6}) to calculate the cosine similarity between $\mathbf{z^{a}}$ and $m^{o}_{i}$, then we obtain the new appearance weight $\bar{\mathbf{w}}^\mathbf{a}$, and finally obtain the appearance features $\mathbf{\hat{z}^{a}}$ without background from $\mathbf{M^{o}}$. This process can be formulated as:

\begin{equation}\label{eq:8}
\mathbf{\hat{z}^{a}}=\bar{\mathbf{w}}^\mathbf{a}\mathbf{M^{o}}= {\textstyle \sum_{i=1}^{N}\bar{w}_{i}^{a}m_{i}^{o}}.
\end{equation}

\begin{figure}[ht]
	\centering
	\includegraphics[width=\linewidth]{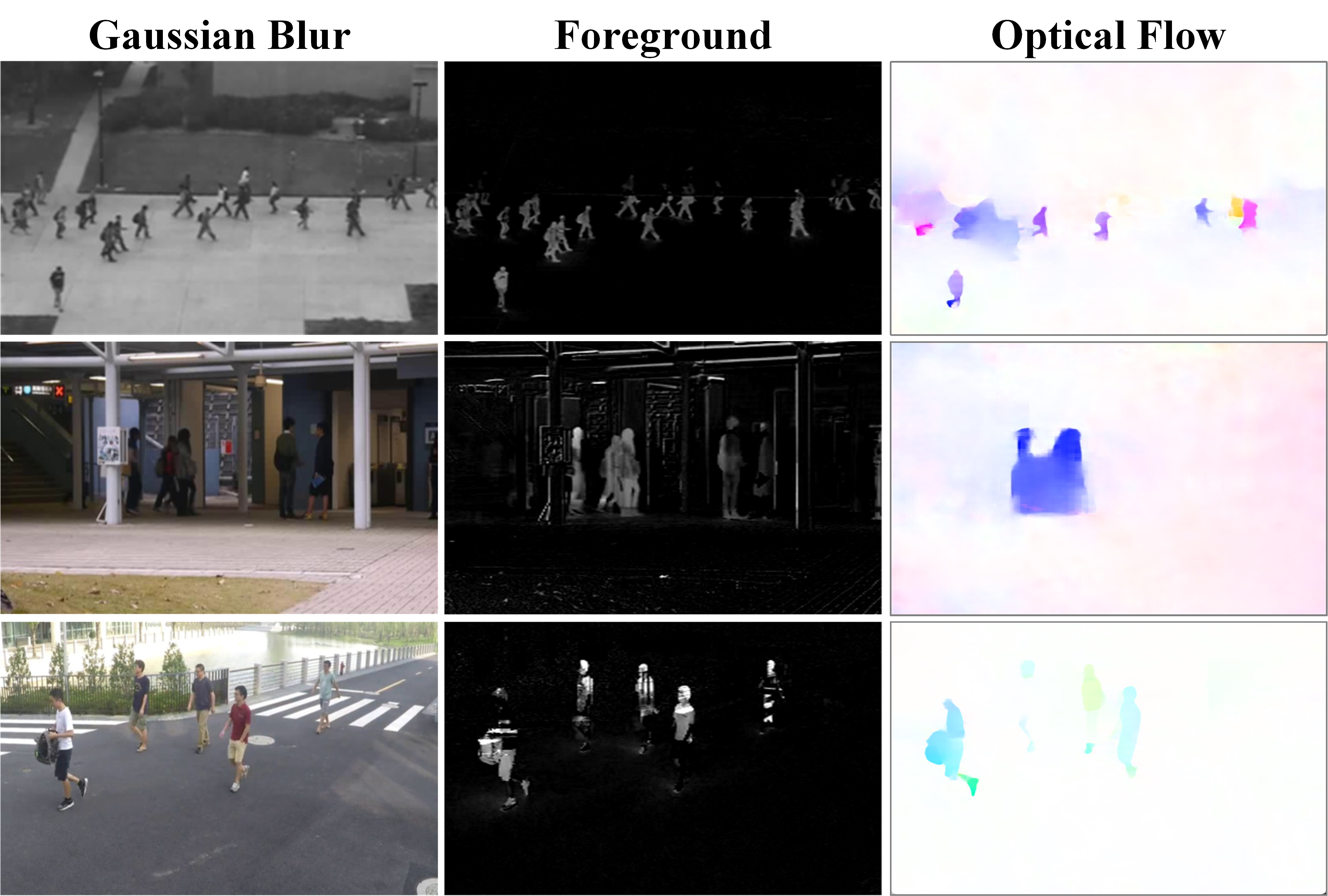}
	\caption{Different input images of three datasets. From top to bottom are UCSD Ped2, CUHK Avenue, and ShanghaiTech.}
	\label{fig:f5}
\end{figure}

We extract two different motion images $O$ by basic foreground extraction and FlowNet2.0 \cite{ilg2017flownet} to design different models for independent experiments. Foreground extraction is subtracting the grayscale frames from the grayscale background frame by frame to obtain the foreground motion images with 3 channels. For the single-scene UCSD Ped2 \cite{sabokrou2015real} and CUHK Avenue \cite{lu2013abnormal} datasets, this method subtracts the global average background, and for the multi-scene ShanghaiTech \cite{luo2017revisit} dataset, it subtracts the average background of the relevant scene. FlowNet2.0 \cite{ilg2017flownet} extracts flow motion images between frames, so the zero convolution of the flow motion encoder needs to be changed to 6 channels. Figure \ref{fig:f5} shows the blurred appearance and two motion images.

\subsection{Loss Function and Anomaly Scoring} \label{sec:sc3.3}

As mentioned above, the proposed method involves the appearance feature, motion feature, and memory module, so we use prediction, gradient, similarity, motion, and compactness losses ($\mathcal{L}_{pred}$, $\mathcal{L}_{grad}$, $\mathcal{L}_{sim}$,  $\mathcal{L}_{motion}$, and $\mathcal{L}_{comp}$, respectively) to train our model:

\begin{equation}\label{eq:9}
\mathcal{L}=\mathcal{L}_{pred}+\mathcal{L}_{grad}+\mathcal{L}_{sim}+\mathcal{L}_{motion}+\mathcal{L}_{comp}.
\end{equation}

For the frame prediction task, we first adopt the mean square error (MSE) to calculate the frame prediction loss:

\begin{equation}\label{eq:10}
\mathcal{L}_{pred}=\left \|I_{t+1}-\hat{I}_{t+1}  \right \| _{2}^{2},
\end{equation}
where $I_{t+1}$ and $\hat{I}_{t+1}$ are the real future frame and the predicted future frame, respectively. Secondly, to improve the pixel reconstruction quality of blurred input images, we deployed the gradient difference loss:

\begin{equation}\label{eq:11}
\begin{gathered}
\mathcal{L}_{grad}=\sum_{i,j} \left \| \left | I_{t}^{i,j}-I_{t}^{i-1,j} \right | -\left | \hat{I}_{t}^{i,j}-\hat{I}_{t}^{i-1,j} \right | \right \|_{1} + \\
\left \| \left | I_{t}^{i,j-1}-I_{t}^{i,j} \right | -\left | \hat{I}_{t}^{i,j-1}-\hat{I}_{t}^{i,j} \right | \right \|_{1},  
\end{gathered}
\end{equation}
where $i$ and $j$ represent the spatial index of the video frame. Then, in the training process the dual-stream encoder, in order to make the appearance features and motion features consistent, we use the similarity loss to minimize the cosine distance between the above two features:

\begin{equation}\label{eq:12}
\mathcal{L}_{sim}=1-\frac{\left \langle f_{a},f_{o} \right \rangle }{\left \| f_{a} \right \|_{2}\left \| f_{o} \right \|_{2} } ,
\end{equation}
where $f_{a}$ and $f_{o}$ represent the final output of the appearance encoder and motion encoder, respectively. Finally, since our method additionally focuses on motion features, we use L1 loss to measure the motion loss between the real future motion $O_{t+1}$ and the predicted future motion $\hat{O}_{t+1}$:

\begin{equation}\label{eq:13}
\mathcal{L}_{motion}=\left \| O_{t+1} -\hat{O}_{t+1} \right \| _{1}.
\end{equation}

To keep a query close to its nearest memory item, we provide a compactness loss that minimizes the MSE between the query and the memory items output:

\begin{equation}\label{eq:14}
\mathcal{L}_{comp}=\left \| \mathbf{z^{o}}-\mathbf{\hat{z}^{o}} \right \| _{2}^{2}.
\end{equation}

In the testing phase, we use the anomaly evaluation proposed by \cite{zhong2022bidirectional} to calculate the Peak Signal-to-Noise Ratio (PSNR) between the real frames $I$ and the predicted frames $\hat{I}$:

\begin{equation}\label{eq:15}
P(I_{t},\hat{I}_{t})=10log_{10}(\frac{1}{\sum_{i=0}^{V}v_{i}}),
\end{equation}
where $v_{i}$ is the patch-based maximum prediction error in scale $i$, and $V$ represents the total number of scales included in the error pyramid. Then, we normalize the PSNR to the $[0, 1]$ range by applying Eq. (\ref{eq:16}), and we apply the Gaussian filter for temporal smoothing over all anomaly scores $S(I_{t})$.

\begin{equation}\label{eq:16}
S(I_{t})=\frac{P(I_{t},\hat{I}_{t})-\text{min}_{t}(P(I_{t},\hat{I}_{t}))}{\text{max}_{t}(P(I_{t},\hat{I}_{t}))-\text{min}_{t}(P(I_{t},\hat{I}_{t}))} .
\end{equation}

\section{EXPERIMENTS}
To demonstrate the effectiveness of the proposed methods, we conduct experiments on three benchmark VAD datasets and compare them with many state-of-the-art methods, and the generalizability of our method is further compared and analyzed with cross-domain methods.

\begin{figure*}[t]
	\centering
	\includegraphics[width=\linewidth]{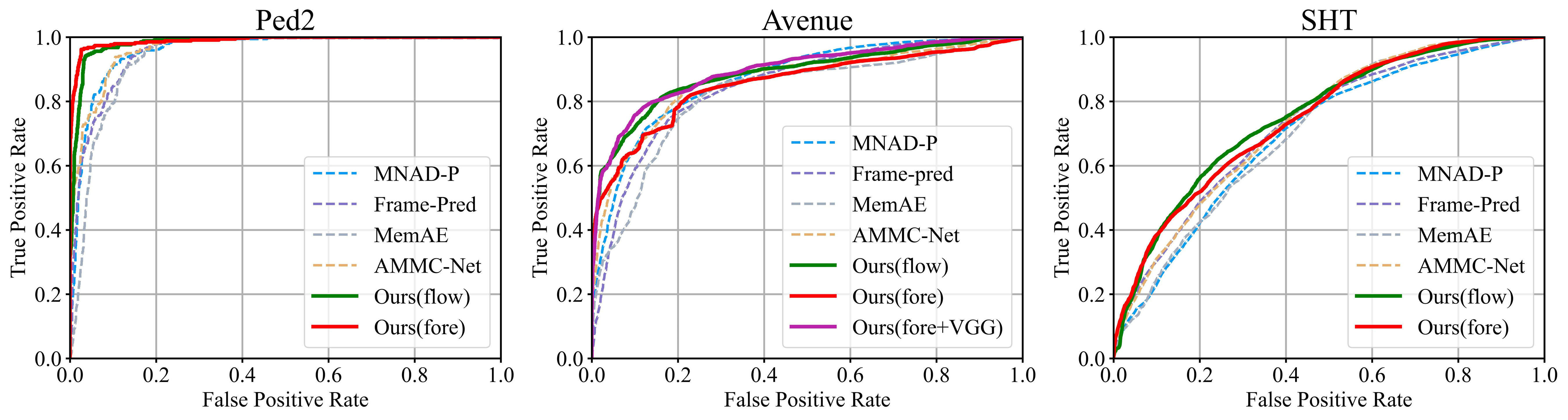}
	\caption{Comparison of frame-level ROC curves on three datasets. The larger area under the curve, the better performance.}
	
	\label{fig:f6}
\end{figure*}

\subsection{Experimental setup} \label{sec:sc4.1}
We conduct experiments on three benchmark datasets UCSD Ped2 \cite{sabokrou2015real}, CUHK Avenue \cite{lu2013abnormal} and ShanghaiTech (SHT) \cite{luo2017revisit}. Table \ref{tab:dataset} summarizes the statistics of these VAD datasets. All the experiments are trained on PyTorch and a single Nvidia RTX 4090 GPU. Our model is trained using the Adam optimizer with a learning rate of 0.0002, and employs cosine annealing scheduler to adjust the learning rate. The number of input frames $t$ is empirically set to 4. All the input images are resized to $256 \times 256$ and normalized the pixel values to the range [-1, 1]. The memory items $N$ are all set to 10. The Gaussian blur hyperparameters $(k, \sigma)$ of Ped2, Avenue, and SHT are set to (3,5), (3,10), and (3,5) respectively. In Ped2, the filter sizes of $W_{1}$ and $W_{2}$ in MRCA are (1,3), (3,5), and (5,7) from the high to the low dimensions, respectively, and are set to (3,3), (3,5) and (3,7) in the Avenue and SHT.

\begin{table}[h]
	\centering
	\caption{The statistics of three unsupervised VAD datasets.}
	\label{tab:dataset}
	
	\begin{tabular}{c|c|c|c|c|c}
		\midrule
		\multirow{2}{*}{Dataset} & \multirow{2}{*}{Scene} & \multicolumn{2}{c|}{Training} & \multicolumn{2}{c}{Testing} \\
		&  & Clips & Frames & Clips & Frames \\
		\midrule
		Ped2   & 1 & 16 & 2,550 & 12 & 2,010 \\
		Avenue & 1& 16 & 15,328 & 21 & 15,324\\
		SHT    & 13& 330 & 274,515 & 107 & 42,883\\
		\midrule
	\end{tabular}
	
\end{table}

\subsection{Comparison with State-of-the-Art methods}
In this section, we compare 19 reconstruction-based (Recon.) \cite{park2020learning,gong2019memorizing,astrid2023pseudobound,wang2023memory,singh2024attention,kommanduri2024dast} and prediction-based (Pred.) \cite{park2020learning,liu2018future,li2023multi,liu2023msn,wang2023video,zhong2022bidirectional,yu2020cloze,cheng2023spatial,yang2023video,yang2022dynamic,cai2021appearance,liang2024c,ijcai2024p96} methods, and conduct a frame-level AUC quantitative comparison to prove the effectiveness of the proposed method, which includes two models: foreground-based model (fore-based) and optical flow-based model (flow-based). The quantitative results are reported in Table \ref{tab:Comparison}, and we show the yielded frame-level ROC curves on different datasets in Figure \ref{fig:f6}.

\begin{table}
	\centering
	\caption{Comparison with state-of-the-art methods on three benchmark datasets. The best and second-best performances are marked in \textbf{bold} and \underline{underlined}, respectively. $^{\circ}$ denotes denoising by VGG16.}
	\label{tab:Comparison}
	\begin{tabular}{c|l|c|c|c}
		\midrule
		& Method    & Ped2 & Avenue & SHT\\ \midrule
		\multirow{6}{*}{\rotatebox{90}{Recon.}} 
		& MemAE \cite{gong2019memorizing}   & 94.1 & 83.3   & 71.2 \\
		& MNAD \cite{park2020learning}     & 90.2 & 82.8   & 69.8 \\
		& PseudoBound \cite{astrid2023pseudobound}   & \underline{98.4} & 87.1   & 73.7\\
		& MAAM-Net \cite{wang2023memory}    & 97.7 & \underline{90.9}   & 71.3 \\
		& DAST-Net \cite{kommanduri2024dast} & 97.9 & 89.8   & 73.7  \\
		& A2D-GAN \cite{singh2024attention}    & 97.4 & \textbf{91.0}   & 74.2 
		\\ \midrule
		\multirow{13}{*}{\rotatebox{90}{Pred.}} 
		& Frame-Pred \cite{liu2018future}    & 95.4 & 85.1   & 72.8 \\
		& MNAD  \cite{park2020learning}      & 97.0 & 88.5   & 70.5  \\
		& VEC \cite{yu2020cloze}    & 97.3 & 89.6   & 74.8 \\
		& AMMC-Net \cite{cai2021appearance}     & 96.6 & 86.6   & 73.7 \\
		& DLAN-AC \cite{yang2022dynamic}       & 97.6 & 89.9   & 74.7 \\
		& MGAN-CL \cite{li2023multi}    & 96.5 & 87.1   & 73.6 \\
		& MSN-net \cite{liu2023msn}     & 97.6 & 89.4   & 73.4 \\
		& STR-VAD \cite{wang2023video}   & \underline{98.4} & 86.1   & 73.2 \\
		& DEDDnet \cite{zhong2022bidirectional}      & 98.1 & 89.0   & 74.5 \\
		& STGCN-FFP \cite{cheng2023spatial}   & 96.9 & 88.4   & 73.7 \\
		& USTN-DSC  \cite{yang2023video}     & 98.1 & 89.9   & 73.8 \\ 
		& C$^{2}$Net  \cite{liang2024c}  & 98.0 & 87.5   & 71.4 \\ 
		& PDM-Net  \cite{ijcai2024p96}    & 97.7 & 88.1   & 74.2 \\ 
		\midrule
		\multirow{2}{*}{\rotatebox{90}{Ours}}
		& Flow-based & \underline{98.4}  & 88.7   & \textbf{75.6} \\
		& Fore-based   & \textbf{99.0} & 85.6 / 89.6$^{\circ}$   & \underline{74.9} \\ 
		\midrule
	\end{tabular}
\end{table}

\noindent\textbf{Quantitative Results.} As shown in Table \ref{tab:Comparison}, our method achieves the best performance on Ped2 and SHT, with AUC reaching at 99.0\% and 75.6\% respectively. Compared with memory modules such as MemAE \cite{gong2019memorizing} and MNAD \cite{park2020learning}, the proposed new memory module better identifies abnormal motion features by focusing on motion features. Conversely, unsatisfying performance for Avenue is caused by the failure to extract clean motion features. This is because the background of Avenue is complex, and we adopt the simplest foreground extraction for efficiency. We have tried to binarize the foreground image using thresholding, similar to paper \cite{zhang2022detecting}, but AUC is sensitive to thresholding. As shown in Figure \ref{fig:f5}, there is severe background noise in the foreground of Avenue, but less noise in the optical flow. If the computation time is flexible, we can refer to the work \cite{wang2021robust}, which utilizes VGG16 to reduce the noise, it can generate less noisy foreground images. The intermediate output features (the second ReLU layer output) of VGG carry a lot of semantic information while filtering out noise. As the output contains redundant channel information, in our model, we thus select the output feature with channel $C=\left \{ 23, 30, 35 \right \}$ as the final denoised foreground images. Eventually, the Avenue is improved by 4.0\% with the cleaner motion information, which again prove the effectiveness of our method. However, the fact that the training efficiency of the fore-based model without VGG is much greater than that of the flow-based model. Despite the fact that it can be observed from the results of \cite{wang2023memory,yu2020cloze} that excellent flow motion features can assist the models in improving the upper-limit, but this comes with an additional computational cost. In contrast, motion images are only used in the training phase of our model, and the testing phase includes only blurred appearance images as input. For example, in Ped2, the model parameters and inference speed (FPS) in the testing phase are 21.98M and 68.5, respectively. Therefore, our model not only improves the detection performance but also ensures the testing efficiency. 

Furthermore, it is noted that A2D-GAN \cite{singh2024attention} and MAAM-Net \cite{wang2023memory} methods achieve good AUC for Avenue. The reasons are: \textbf{(1)} both methods use two weighted anomaly scores for the final result, which are more effective than a single frame score when facing complex environments. However, finding a universal weighting strategy is challenging; \textbf{(2)} A2D-GAN includes both image and noise discriminators, with the noise discriminator enhancing robustness but increasing training difficulty; \textbf{(3)} MAAM-Net employs two tasks, frame reconstruction and optical flow prediction, which improves anomaly sensitivity to subtle errors based on local patches. However, since patches are based on the average size of the person and there is significant variance in person size in SHT, AUC suffers. 


\begin{figure*}
	\centering
	\includegraphics[width=\linewidth]{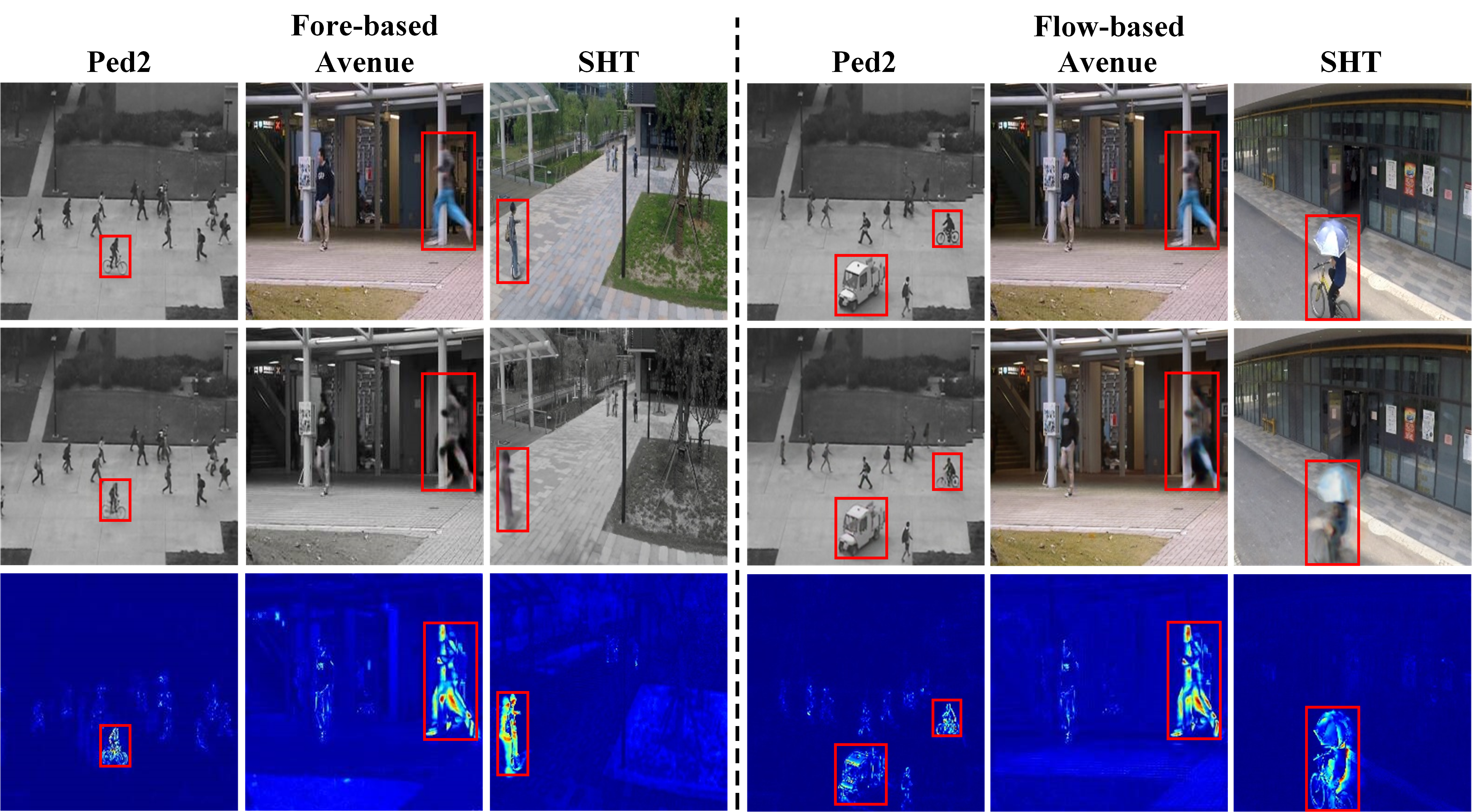}
	\caption{Visualization of frame prediction. From top to bottom, we show real future frames, predicted future frames, and error maps. In the error map, brighter color indicates larger errors. The objects remarked with red borders are the anomaly.}
	\label{fig:f7}
\end{figure*}

\noindent\textbf{Qualitative Results.} To demonstrate the visual detection effect of the proposed method, we visualize representative video frames in Figure \ref{fig:f7}. Several phenomena can be intuitively observed through Figure \ref{fig:f7}: \textbf{(1)} With color images, the predicted frame of the fore-based model is the grayscale image, while the flow-based model remains in color. This is because we have extracted the foreground motion features from grayscale images, and grayscale features can influence the model's perception of color. This situation also occurs in \cite{zhong2022cascade} due to the black mask patch pseudo-anomaly. \textbf{(2)} The abnormal samples in predicted frames have obvious blur. In contrast, our model removes the blur features from the normal samples, because attention can focus only on the normal samples. \textbf{(3)} We visualize the same video frames on Avenue and observe that the fore-based model preserves appearance errors against anomalies in addition to producing color errors against anomalies, which is also observed in low-saturation color video images in the SHT.

\begin{figure}
	\centering
	\includegraphics[width=\linewidth]{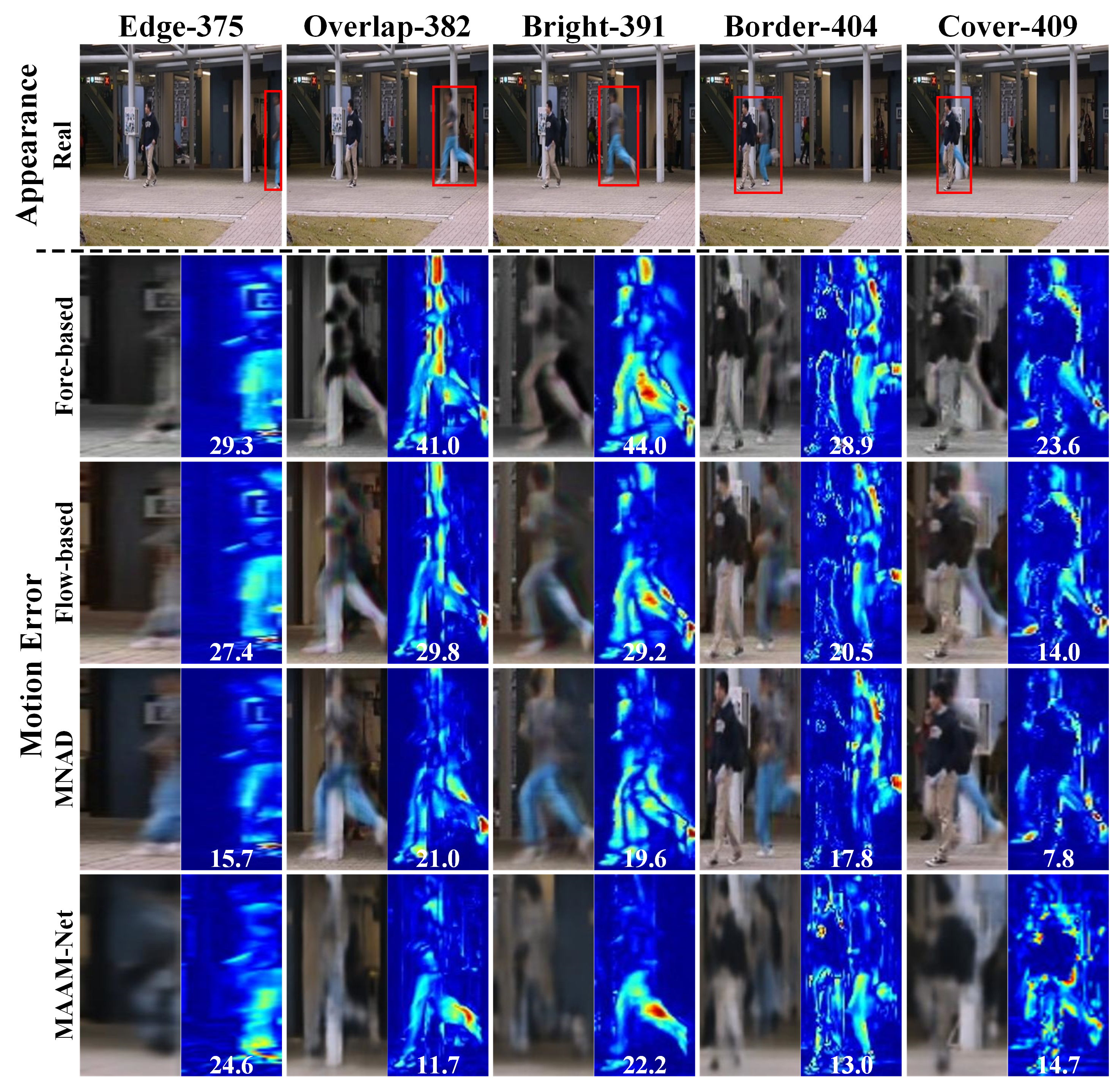}
	\caption{The anomaly motion error in different states. Each column in the error includes prediction/reconstruction details and error details. The numbers denote the sum-square error between the real anomaly and the predicted anomaly.}
	\label{fig:f8}
\end{figure}

Furthermore, to illustrate the special attention to the motion features of our method, we visualize and compare the motion details in the anomaly with two memory module methods, MNAD (Pred.) \cite{park2020learning} and MAAM-Net (Recon.) \cite{wang2023memory}. Figure \ref{fig:f8} illustrates the five states in the consecutive frames, including the frame edge (Edge), foreground overlap (Overlap), bright background (Bright), normal and abnormal border (Border), and anomaly covered (Cover). By Figure \ref{fig:f8}, it can be noted: \textbf{(1)} The proposed method predicts anomaly in five states and has the most pronounced errors compared to the other methods. In Border, the weak errors in the normal sample are represented in the object contour, which is also observed in most VAD methods. \textbf{(2)} In the Edge and Bright, MNAD performs poorly due to delayed updates of memory items to affect detection. \textbf{(3)} In Overlap and Cover, our memory module focuses on the normal motion, thus maximally ignoring the recovery from anomalous motion, and other methods are not able to efficiently distinguish these two states. \textbf{(4)} In the Border and the Cover,  MAAM-Net can be found that it cannot separate normal from abnormal, and produce significant errors. The sum square errors are used to measure the quality of the generated anomaly, the higher indicates that the anomaly is recognized more obviously. For Border, the error result only calculates the anomaly part. The normal error in Cover cannot be avoided, so MAAM-Net should be lower than 14.7.

\subsection{Curve of Anomaly Score}

\begin{figure*}
	\centering
	\includegraphics[width=\linewidth]{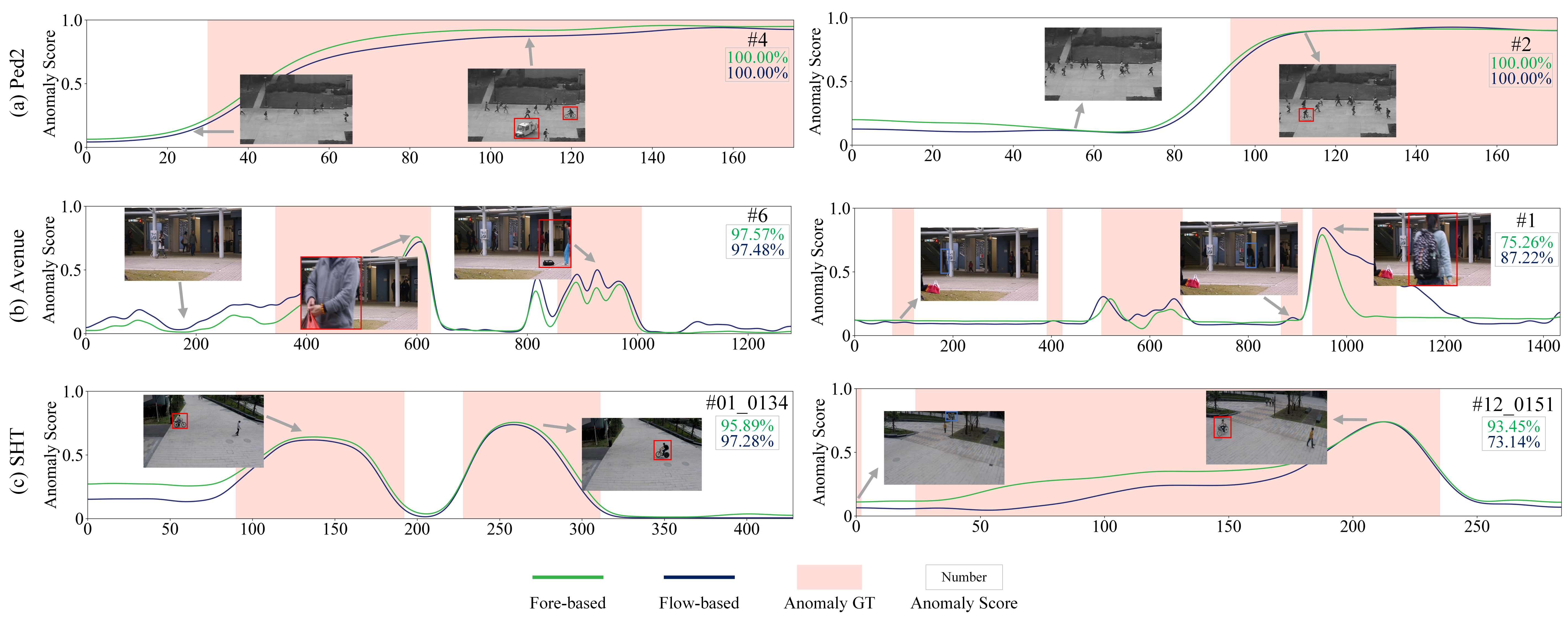}
	\caption{Examples of frame-level anomaly curves of test videos on (a) Ped2, (b) Avenue and (c) SHT. The \textcolor{green}{green} and {\color[RGB]{31,56,100}{navy}} indicate the results of the fore-based and the flow-based model, respectively. The red (blue) bounding boxes represent the detected (undetected) objects, and the red regions indicate the anomaly ground truth. The lines and accuracy numbers are anomaly curves and AUCs for the corresponding scenes, respectively. Zoom in better.}
	\label{fig:score}
\end{figure*}

This section further visualizes the anomaly detection curves for some of the test clips in the three datasets. \textbf{(1)} We achieve perfect frame-level AUCs for both models in test video \#2 and \#4 in Ped2, as shown in Figure \ref{fig:score}(a). The curves show that the model clearly distinguishes between normal and abnormal. This is because Ped2 is the single scene dataset composed of grayscale images, which enables models to extract motion features better and thus achieve higher performance. \textbf{(2)} In the previous section, we mentioned that the reason for the poor performance in Avenue is due to poor extraction of motion features, for instance, in the blue bounding box in test video \#1, as shown in Figure \ref{fig:score}(b). The anomaly motion is similar to the background, which causes motion feature extraction to be hard, thus the model is unable to determine the anomaly effectively. But as shown in test video \#6, with foreground motion and optical flow motion extraction unaffected, our model gets better performance. \textbf{(3)} Compared with Ped2 and Avenue, there are more scenes and more anomalies in the ShanghaiTech dataset, as shown in test video \#01\_0134 at Figure \ref{fig:score}(c). Our method can discriminate multiple anomalies in the same scene and efficiently discriminate anomalies with different orientations and appearances. Besides, the inter-domain differences in different scenes lead to insensitivity to some anomalies, as shown in the blue bounding box in test video \#12\_0151, which cannot be efficiently detected for the anomaly that is occluded and at the edge of the surveillance video.

\subsection{Cross-dataset Validation}
To demonstrate the generalizability of our proposed method, we conduct the cross-dataset validation comparison with few-shot methods \cite{lu2020few,wang2022few,huang2022boosting,lv2021learning,zhang2024cognition} and zero-shot methods \cite{georgescu2021background,aich2023cross}. Following \cite{lu2020few}, we use  SHT as the training dataset and do not use small datasets Ped2/Avenue as the training dataset, because they are unable to draw reasonable conclusions. The SHT contains up to 300,000 video frames and more than 150 types of anomalous events, which far exceeds the size and anomaly variety of the testing datasets Ped2 and Avenue. The multi-scene and inconsistent-size targets of SHT provide data diversity and generalization for cross-domain detection. Table \ref{tab:CROSS} shows the cross-dataset performance trained on the large dataset SHT and tested on Ped2/Avenue.

\begin{table}[h]
	\centering
	\caption{Comparison with cross-domain methods only trained on SHT. The best performance in each category is highlighted in \textbf{bold}. $^{+5}$: 5-shot.  $^{+10}$: 10-shot.}
	\label{tab:CROSS}
	
	\begin{tabular}{c|c|c|c}
		\midrule
		& Method & Ped2 & Avenue\\
		\midrule
		\multirow{5}{*}{\rotatebox{90}{Few-shot}}  & r-GAN \cite{lu2020few}& 92.80$^{+10}$  & 78.79$^{+10}$ \\
		& MPN \cite{lv2021learning}& 95.75$^{+10}$ & 81.69$^{+10}$  \\
		& Wang et al. \cite{wang2022few} & 92.45$^{+10}$ & 84.46$^{+10}$   \\
		& VADNet \cite{huang2022boosting} & 95.12$^{+10}$  & 82.62$^{+10}$  \\
		& CG-VAD \cite{zhang2024cognition} & \textbf{96.70}$^{+5}$	 & \textbf{85.10}$^{+5}$  \\
		\midrule
		\multirow{8}{*}{\rotatebox{90}{Zero-shot}} & BG-Agnostic \cite{georgescu2021background} & 90.60 & 83.60 \\
		
		& zxVAD (SHT) \cite{aich2023cross} & 95.78	 & 82.28 \\
		& zxVAD (HMDB) \cite{aich2023cross} & 95.74	 & 83.19 \\
		& zxVAD (UCF101) \cite{aich2023cross} & 95.80	 & 82.25 \\
		& zxVAD (Jester) \cite{aich2023cross} & 95.62	 & 82.49 \\
		& Ada-VAD \cite{guo2024ada} & \textbf{97.53}	 & 83.62 \\
		& Fore-based (Ours) & 95.36 & 83.36 \\
		& Flow-based (Ours) & \textbf{97.53} &  \textbf{87.47} \\
		\midrule
	\end{tabular}
	
\end{table}

The performance of few-shot methods depends on the fine-tuning with different numbers of \textbf{K-shot}, and the optimal performance is all obtained under \textbf{K=10} or \textbf{K=5}, while the performance of these methods degrades under \textbf{K=0}, for instance in CG-VAD \cite{zhang2024cognition}, drop to 92.90\% (-3.80\%) and 79.80\% (-5.30\%), respectively. Zero-shot methods do not require additional target domains for fine-tuning. Similar to our method, zxVAD \cite{aich2023cross} employs different auxiliary data (SHT, HMDB, UCF101, and Jester) as pseudo-anomaly, but the auxiliary data inevitably increases the requirement of the computation time and memory. In contrast, our method achieves cross-domain adaptive by focusing on the motion features and eliminating background interference. It can be seen that our method has achieved very impressive performance, especially achieving SOTA on Ped2 and Avenue.

\subsection{Ablation Studies}

\noindent\textbf{Component Analysis.} To elucidate the contribution of each component to the proposed method, we design the component ablation study of the fore-based model, as shown in Table \ref{tab:ABLATION}. Firstly, the removal of the memory module resulted in the most significant performance degradation, indicating that the memory module enhances the ability to represent normal motions. This further illustrates the special focus on motion features in the new memory module, which enables more explicit retrieval and separation of normal and abnormal features by memory items. Secondly, we use blur-free images as input, and although the MRCA still pays attention to normally distributed features, this increases the risk of over-fitting and reduces the focus on anomalous information during the testing process. Then, we remove the MRCA under blurred appearance images, which undoubtedly cannot prevent anomalies from entering the decoder, with its impact being second only to the removal of the memory module. Finally, we remove the zero convolution layer in the motion encoder. Although this had the least impact compared to other variant models, the performance of learning from two types of motion features suggests that the cleaner features, the more likely they are to achieve better performance. Therefore, zero convolution is necessary to eliminate harmful noise and ensure higher accuracy. Naturally, to further demonstrate the effectiveness of the deblurring solution for VAD, we exclude the motion image branch (w/o Motion), and the results demonstrate that better performance is still obtained in the configuration of inputting only blurred appearance images. Especially, it can be demonstrated by Avenue that the quality of motion images also impacts performance.

\begin{table}
	\centering
	\caption{Component ablation study of the fore-based model.}
	\label{tab:ABLATION}
	\setlength{\tabcolsep}{1.7mm}{
		\begin{tabular}{c|c|c|c|c|c}
			\toprule
			\multirow{2}{*}{\diagbox[width=5em,trim=l]{Dataset}{Model}} & {w/o}& {w/o}& {w/o}& {w/o}& {w/o}  \\ 
			& Memory& Blur& MRCA& Zero & Motion  \\
			\midrule
			Ped2   & 97.1& 97.7& 97.4& 98.3& 98.1 \\
			Avenue & 83.2& 83.5& 83.5& 83.7& 86.6 \\
			SHT    & 73.5& 73.7& 73.5& 73.6& 74.9 \\
			\bottomrule
		\end{tabular}
	}
\end{table}

\begin{figure}[h]
	\centering
	\includegraphics[width=1.1\linewidth]{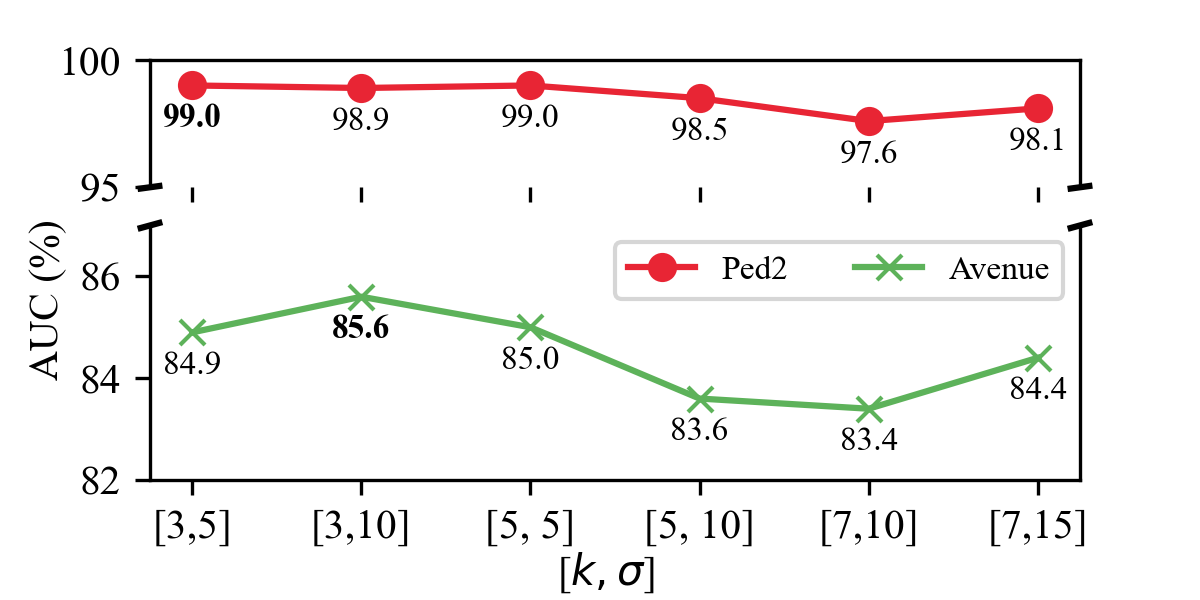}
	\caption{Evaluate the six settings of blur hyperparameter and test on Ped2 and Avenue. }
	\label{fig:blur}
\end{figure}

\noindent\textbf{Blur Hyperparameter.} We show the results of the blur hyperparameter ablation experiment based on the fore-based model in Figure \ref{fig:blur}. It can be seen that the model is almost unaffected in Ped2 by the different blur hyperparameters, the performance varies within a reasonable range. The average of the six settings is about 98.4\%, which maintains good performance. On the other hand, due to the negative factors of the object's own unclarity and illumination variations in Avenue, too large a degree of blur will further affect the discrimination and thus hurt the accuracy. But the maximum difference AUC between the six settings is also only 2.2\%, proving once again the robustness of our method.

\begin{figure}[h]
	\centering
	\includegraphics[width=\linewidth]{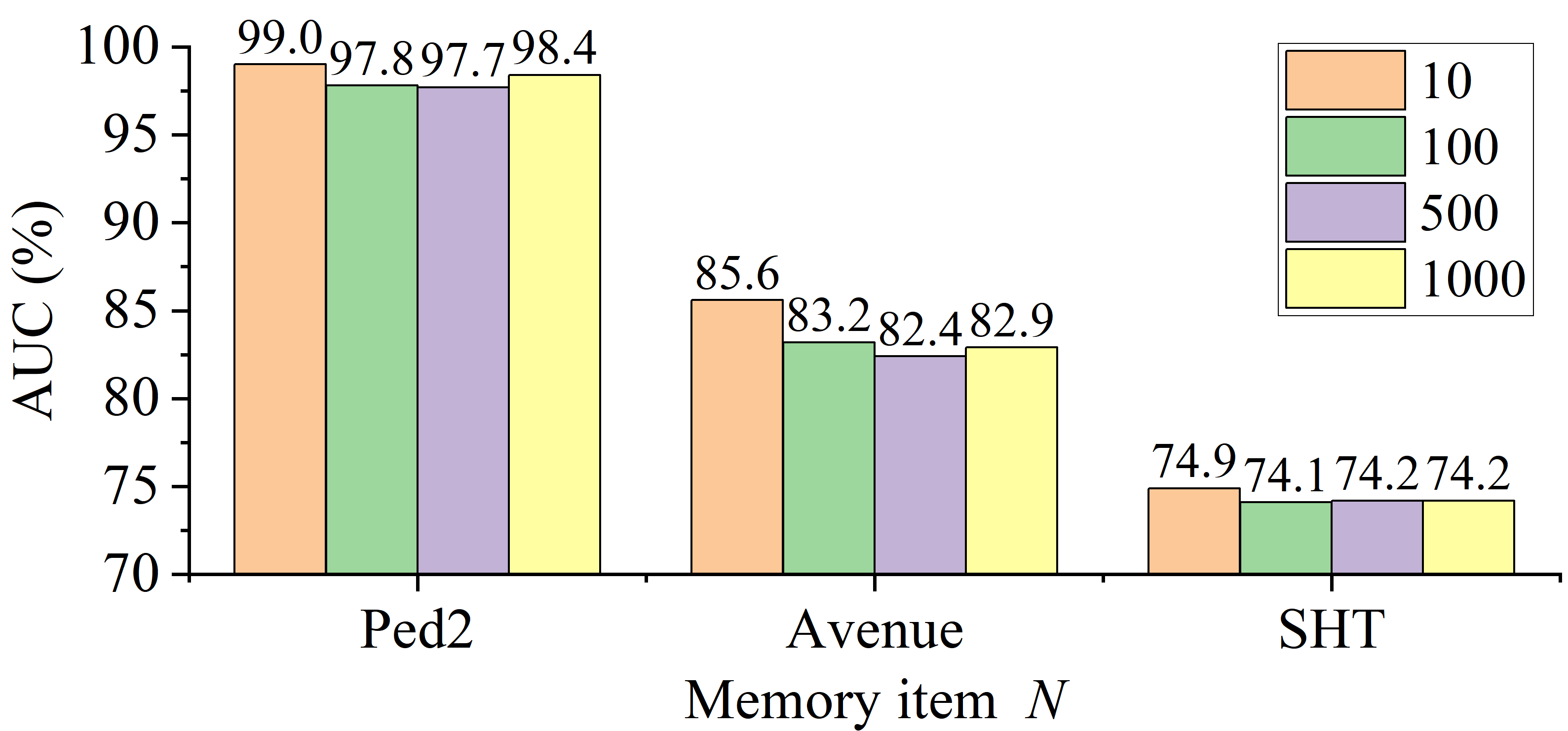}
	\caption{Different memory item settings in Ped2 and Avenue. }
	\label{fig:memory}
\end{figure}

\noindent\textbf{Memory item.} Most memory models \cite{gong2019memorizing,cai2021appearance} rely on many memory items to record the distribution of normal features, whereas we use tiny memory sizes to record the motion features and mitigate undesirable generalizations. Figure \ref{fig:memory} shows the ablation results based on the fore-based model for different numbers of memory items. It can noted that the increase in the number of items decreases the performance and leads to overfitting. Despite the different performance of AUC at $N=1000$, it is still much lower than the performance at $N=10$. Thus causing too many memory items can increase the sensitivity of similarity weights to anomalous retrievals during testing.

\section{CONCLUSION}

We propose a novel VAD method from the perspective of generated anomaly representation and motion features. We transform VAD into the deblurring process for normal samples, ignoring the blurred real anomaly through attention, and design a new memory module to focus on motion features that may contain anomalies, thereby overcoming the over-fitting by background interference and the cross-domain VAD problem. Extensive experimental evaluations on three benchmark datasets demonstrate that our model outperforms the state-of-the-art methods. Our performance demonstrates the potential for focusing on motion-related samples with background-agnostic to become a favorable direction for addressing the VAD problem. Most importantly, our method achieves desirable performance in cross-dataset validation and provides a simple idea for future domain-adaptive models.


\section*{Acknowledgement}

This research was funded by Natural Science Foundation of Shaanxi Province, China(2024JC-ZDXM-35, 2024JC-YBMS-458, 2024JC-YBMS-573), National Natural Science Foundation of China (No.52275511) and Young Talent Fund of Association for Science and Technology in Shaanxi, China(20240146).


{\small
\bibliographystyle{cvm}
\bibliography{references}
}

\end{document}